\documentclass[letterpaper, 10 pt, conference]{ieeeconf}  %

\IEEEoverridecommandlockouts                              %

\usepackage{graphics} %
\usepackage{epsfig} %
\usepackage{mathptmx} %
\usepackage{times} %
\usepackage{amsmath} %
\usepackage{amssymb}  %
\usepackage{balance}
\usepackage{booktabs}
\usepackage{multirow}

\title{\LARGE \bf
RoboMMT: Learning Stable Tracking of Modulated\\Expressive and Interactive Robot Motion
}

\author{Till Hielscher and Kai O. Arras%
\thanks{T. Hielscher and K. O. Arras are with the Socially Intelligent Robotics Lab, Institute for Artificial Intelligence, University of Stuttgart, Germany {\tt\small till.hielscher@ki.uni-stuttgart.de}}%
\thanks{The authors thank the International Max Planck Research School for Intelligent Systems (IMPRS-IS) for supporting Till Hielscher.}
}

\begin{document}

\maketitle
\thispagestyle{empty}
\pagestyle{empty}

\begin{abstract}
Expressive robot motion can improve naturalness, effectiveness, and perceived likability in human-robot interaction. When deployed on dynamically unstable platforms, however, reactive execution of expressive body motion may bring the robot to its stability limits, especially when expressivity is decoupled from physical control. In this paper, we present RoboMMT (Robot Modulated Motion Tracking), a modular control architecture that couples parameterized expressive motion synthesis, physics-based character control, and reactive adaptation to spatial interaction goals. Our approach trains a reference-conditioned tracking policy across a broad domain of expressivity, enabling it to faithfully realize the commanded motions while maintaining dynamic stability. Goal reactivity is achieved by training lightweight, task-specific interaction policies that modulate the reference generator's spatial goal-attraction point, with the stabilizing tracking policy kept in the loop. Evaluations on a quadruped and two humanoid robots show that the learned tracker remains stable and accurate across the entire modulation domain, whereas trackers trained without this coverage lose balance, tracking accuracy, or both. We show that layering closed-loop goal adaptation over this foundation enables interactive facing and pointing behaviors, while maintaining stable execution and comparable reference-tracking fidelity. The robot learns to proactively redistribute its weight and take corrective steps. Zero-shot sim-to-real transfer is demonstrated through real-robot experiments on a quadruped under heavy disturbances.
\end{abstract}

\section{INTRODUCTION}
    Robots operating in human environments require both verbal and non-verbal channels for social communication, particularly through body language. Beyond executing their primary tasks, these robots must convey social signals to communicate intent or affect. Depending on the context, they need mechanisms to stylize their body language by modulating timing, extent, or motion dynamics in response to how the interaction unfolds \cite{saerbeck2010perception,venture2019robot,cha2018survey}. Here, we address the challenge of jointly executing expressive body language while maintaining stability and physical feasibility under complex platform dynamics.
    
\begin{figure}
        \centering\includegraphics[width=0.95\linewidth]{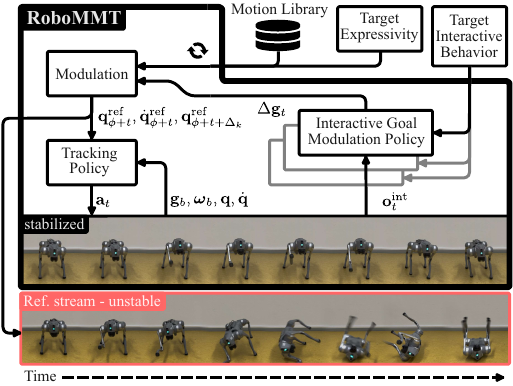}
             
        \caption{We address the task of executing expressive body language for social interaction while maintaining physical feasibility under complex platform dynamics. The figure shows our approach, \textit{RoboMMT}, which learns tracking and goal-modulation policies that keep the robot stable while performing a modulated greeting gesture. Without it, the robot tips over.}
        \label{fig:overview}
    \end{figure}

    Recent advances in character animation use physics-based control through reinforcement learning to enable robots to reproduce complex, expressive reference motions while accounting for modeled system dynamics \cite{peng2018deepmimic,serifi2024vmp}. This way, behaviors authored for virtual characters can be realized on physical platforms \cite{grandia2024design,mueller2025olaf}.    
    Other lines of work propose motion representations that make stylization directly and interpretably controllable. Rather than specifying expressivity through authored trajectories, recent kinematic approaches represent expressive characteristics along interpretable dimensions, enabling the design of specific motion styles using Laban features \cite{knight2015layering} or Disney's animation principles \cite{hielscher2025interactive}. Such representations can then be embedded into continuous control systems that generate interactive body language automatically from multimodal context inputs \cite{hielscher2026context}. However, expressive reference generation and physical execution remain largely decoupled \cite{bergamin2019drecon}, often rendering online expressivity adjustments physically unfeasible or unstable (see Fig.~\ref{fig:overview}).

    To bridge this gap, we present an approach that builds on parametrized motion synthesis with physics-based control and reactive interactive behavior. We ground motion expressivity in the Dynamic Movement Primitive (DMP) formulation of parametrized animation principles of prior work \cite{hielscher2025interactive, hielscher2026context} and train a policy to track continuously modulated references while maintaining physical stability. A decoupled interaction loop, driven by interchangeable, lightweight policies, reacts to interaction goals (a person or object in space) by adapting the spatial direction of the expressive reference stream, which is then stabilized by the tracking policy. Specifically, our contributions are:
    \begin{itemize}
        \item A method for stable execution of continuously modulated reference motions for social human-robot interaction that trains the tracking policy over the DMP-induced reference domain for robust execution across expressive variations. 
        \item An interaction layer that modulates the goal attraction point of the reference generating DMP with lightweight plug-and-play policies to perform spatially directed execution with the stabilizing tracking policy in the loop.
    \end{itemize}
        Experiments on multiple robot platforms in simulation show the method's ability to perform modulated motion with high stability and low tracking errors. A sim-to-real deployment of the system on a quadruped robot demonstrates interaction sequences and disturbance stability.

        Note that this paper does not raise claims related to the way how expressivity is generated or perceived, for which a user study would be required. Instead, the paper makes the technical contribution of learning to dynamically execute expressive body motion evaluated against the respective task and tracking metrics.

\section{RELATED WORK}
\label{sec:rw}

    \subsection{Expressive and Interactive Robot Motion}
    \label{sec:rw_expressive}
     
        Generating expressive robot motion across humanoid and non-humanoid systems typically relies on manual keyframing, classic animation heuristics, or retargeted motion capture \cite{thomas1981illusion,ribeiro2012illusion,saerbeck2010perception,venture2019robot,cha2018survey}. Despite communicating affect and intent effectively, these authoring pipelines remain inherently bottlenecked, where every individual expressive behavior requires dedicated, time-consuming manual specification.
         
        Parameterized motion representations capture expressive variation through a reduced set of variables. Laban Effort features can specify expressive qualities while preserving task motion, and related approaches parameterize kinematic and temporal properties like velocity, spatial extent, and timing \cite{knight2014expressive,knight2015layering,xu2013mood,zhou2017expressive}. Beyond parameterizing descriptive traits, representations built on mathematically integrated animation principles allow systems to automatically and interactively modulate expressive motion \cite{hielscher2025interactive}. Learning-based methods offer another avenue by extracting latent style spaces directly from motion datasets \cite{sripathy2022teaching}. Although these methods vary in their underlying structures, each uses parameterized expressivity to produce a variety of styles. Compared to latent representations, explicit parameterization keeps variables interpretable so designers and high-level controllers can adjust them directly.
         
        Interactive robots must also couple this expressivity to the dynamic state of nearby humans. Social and movement-based platforms typically achieve this by directing attention through gaze or body orientation \cite{admoni2017social}, regulating interpersonal distance and spatial layout \cite{mead2017autonomous}, and adapting action timing to human behavior \cite{huang2016anticipatory}. These capabilities highlight how responsiveness and behavioral contingency influence the perceived quality of interaction \cite{hoffman2014responsiveness,kozima2009keepon}. Recent systems build on these insights by explicitly adapting behavior based on human observations, which includes generating interaction-driven control commands \cite{christen2025autonomous} and using context to select and modulate expressive movement parameters \cite{hielscher2026context}.
         
        However, these interactive and parameter-driven methods predominantly operate at the kinematic level. Rapidly modulating expressive references online provides no guarantees of physical feasibility on dynamic platforms, often inducing balance loss or actuator saturation. Bridging this gap requires decoupling interactive parameter adjustment from low-level physical tracking.
     
    \subsection{Physics-Based Character Animation and Robotic Characters}
    \label{sec:rw_physics}
    
        Physics-based character animation translates kinematically authored trajectories into physical execution by learning policies that imitate reference motions under simulated dynamics. Deep reinforcement learning has advanced these capabilities through explicit tracking rewards \cite{peng2018deepmimic}, adversarial motion priors \cite{peng2021amp}, reusable skill embeddings \cite{peng2022ase}, and masked reference completion \cite{tessler2024maskedmimic}. Hierarchical two-layer architectures further decouple high-level reference generation from low-level physical control \cite{bergamin2019drecon}, drawing from large-scale motion capture datasets to supply target behaviors \cite{mahmood2019amass}.
    
        While dataset-driven approaches thrive on humanoid platforms that closely resemble human morphology \cite{cheng2024expressive,he2024h2o,serifi2024vmp}, non-humanoid robotic characters lack comparable motion repositories. Consequently, research on non-humanoid characters emphasizes morphology-independent policy learning, imitation control, and physics-aware retargeting across diverse robot structures \cite{peng2018deepmimic,mueller2026reactor}. On these platforms, carefully authored physical traits and reference trajectories define the character's expressivity and identity \cite{grandia2024design,mueller2025olaf}, leaving the learned controller to realize these predefined behaviors under real-world dynamics.
    
        More recently, runtime adaptation has been introduced directly into physics-based controllers. For instance, Hopkins et al. \cite{hopkins2024interactive} expose continuous style controls for interactive gait design, Alegre et al. \cite{alegre2025amor} condition tracking controllers on shifting multi-objective preferences at inference, and Kobayashi et al. \cite{kobayashi2026emolo} inject handcrafted emotion parameters into locomotion policies.
    
        While these developments show that stable physics-based controllers can accommodate behavioral variations beyond a single static motion, current adaptation mechanisms remain bound to uninterpretable latent spaces and scalar preference weights, or narrow locomotion styles. In contrast, our work grounds controller execution in a broad, interpretable DMP parameter space, coupling stable physical tracking with closed-loop goal adaptation for interactive robotic characters.

\begin{figure*}
\vspace{0.3cm}
    \centering
    \includegraphics[width=0.99\linewidth]{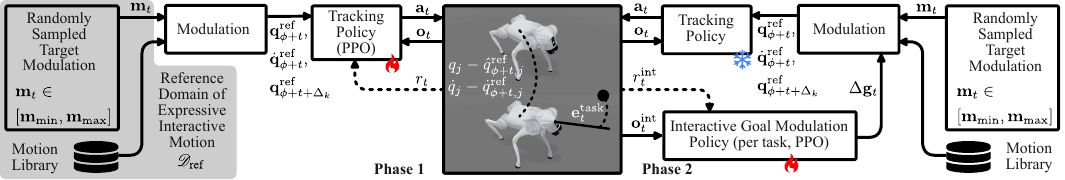}
             \vspace{-0.3cm}
    \caption{Training is organized in two phases in a shared parallelized physics simulation environment. In Phase 1, a policy learns to track continuous reference motion covering the domain of modulated expressivity $\mathcal{D}_{\mathrm{ref}}$, by reducing imitation errors. In Phase 2 (right), the tracking policy is frozen and acts in the loop to train lightweight interaction policies that modulate the reference generators' spatial goal attraction point to reduce the task-specific error.}
    \label{fig:training}
\end{figure*}

\section{BACKGROUND}

    \subsection{Expressive Motion Representation}
    \label{sec:bg_dmp}

        We adopt the parametrized motion representation from~\cite{hielscher2025interactive}, which formulates classical animation principles~\cite{thomas1981illusion} as Dynamic Movement Primitives (DMPs)~\cite{ijspeert2013dynamical}. A DMP defines a multi-dimensional joint trajectory $y$ using a second-order dynamical system driven by a phase $x$ toward a goal state $g$:
        \begin{equation}
            \tau^2 \ddot{y} = \alpha \left( \beta (g - y) - \tau \dot{y} \right) + f(x),
            \label{eq:dmp}
        \end{equation}
        where execution time $\tau$ scales temporal progression, gains $\alpha = 4\beta$ guarantee convergence to $y = g, \dot{y} = 0$, and weights $w$ of the nonlinear forcing term $f(x)$ are fitted from demonstrated trajectories using locally weighted regression.
        
        Expressivity is controlled by adjusting different system components through a low-dimensional set of interpretable parameters, mapped directly to individual animation principles. Trajectory features are altered by filtering or noise-perturbing forcing weights $w$ (Arcs $p_{\text{Arcs}}$, Variability $p_{\text{Rand}}$), scaling $f(x)$ (Exaggeration $p_{\text{Exa}}$), or inverting initial acceleration (Anticipation $p_{\text{Ant}}$). Temporal profile is shaped by scaling $\tau$ (Timing $p_{\text{Time}}$) and reshaping the phase function (Slow In Slow Out), whereas inter-joint coupling models physical connectivity (Follow Through $p_{\text{Follow}}$) and modulating $g$ shifts spatial intent ($p_{\text{Goal}}$). This parameterization preserves the core identity of the authored motion while offering intuitive stylistic control.
        
        Because DMPs are stable attractor systems, library primitives can be smoothly concatenated into a continuous reference stream~\cite{hielscher2026context}. By aggregating the modulation parameters into a bounded vector $\mathbf{m}_t \in [\mathbf{m}_{\min}, \mathbf{m}_{\max}]$ updated at control frequency, integration of~\eqref{eq:dmp} across all dimensions yields
        \begin{equation}
            \mathbf{q}^{\mathrm{ref}}_t \equiv \mathbf{q}^{\mathrm{ref}}_t\!\left( \mathbf{m}_t\right),
            \label{eq:reference_stream}
        \end{equation}
        enabling online adaptation of motion expressivity during execution without trajectory replanning.

    \subsection{Reference-Conditioned Reinforcement Learning}
    \label{sec:bg_tracking}
    
        Reinforcement learning (RL) formulates continuous decision-making as a Markov Decision Process (MDP) $(\mathcal{S}, \mathcal{A}, \mathcal{P}, r, \gamma)$, parameterized by state space $\mathcal{S}$, action space $\mathcal{A}$, transition dynamics $\mathcal{P}(\mathbf{s}_{t+1} \mid \mathbf{s}_t, \mathbf{a}_t)$, reward function $r(\mathbf{s}_t, \mathbf{a}_t)$, and discount factor $\gamma \in [0, 1)$. The objective is to learn the policy $\pi_\theta$ to maximize expected discounted return $J(\theta) = \mathbb{E}_{\pi_\theta} [\sum_{t=0}^{\infty} \gamma^t r_t]$ using PPO~\cite{schulman2017ppo}.        
        
        To track dynamic targets rather than solve static tasks, the framework extends to reference-conditioned control~\cite{schaul2015universal} by factoring the state $\mathbf{s}_t \in \mathcal{S}$ into a physical state $\mathbf{s}_t^{\mathrm{phys}}$ and a target reference specification $\boldsymbol{\rho}_t$: $\mathbf{s}_t = \left( \mathbf{s}_t^{\mathrm{phys}}, \, \boldsymbol{\rho}_t \right)$. The descriptor $\boldsymbol{\rho}_t$ encodes target behaviors, such as instantaneous configurations, lookahead trajectories, or motion parameters. A physical agent accesses this state through an observation $\mathbf{o}_t = (\mathbf{o}_t^{\mathrm{phys}}, \boldsymbol{\rho}_t^{\mathrm{obs}})$ that inherits the same factorization into sensed physical quantities and commanded references. The policy $\pi_\theta(\mathbf{a}_t \mid \mathbf{o}_t)$ drives physical state transitions $\mathbf{s}_{t+1}^{\mathrm{phys}} \sim \mathcal{P}^{\mathrm{phys}}(\cdot \mid \mathbf{s}_t^{\mathrm{phys}}, \mathbf{a}_t)$ toward $\boldsymbol{\rho}_t$, while $r$ measures tracking precision subject to physical stability and control effort constraints.

\section{METHOD OVERVIEW}
\label{sec:overview}

    Execution of interactive body language requires a control architecture that preserves dynamic balance without compromising expressive motion style. As discussed in Sec.~\ref{sec:rw}, existing kinematic motion generators enable continuous style modulation but lack physical awareness, whereas traditional tracking controllers maintain balance by smoothing out stylized nuances. We bridge this gap through a three-stage hierarchical pipeline (Fig.~\ref{fig:overview}). First, a DMP-based reference stream generator to produce continuous motion with interpretable expressive variation (Sec.~\ref{sec:bg_dmp}). Second, a single reference-conditioned policy that executes these trajectories under complex platform dynamics, retaining the kinematic characteristics encoded by the modulated reference (Sec.~\ref{sec:generalized_tracking}). Finally, closed-loop goal adaptation modules that dynamically update task targets, enabling real-time reactivity without destabilizing low-level balance (Sec.~\ref{sec:goal_modulation}). %
    Throughout, we assume that expressive motion, including its communicative purpose and its generation, is an input to our method, and not a learning objective.
    
    Training proceeds in two phases (Fig.~\ref{fig:training}). In Phase 1, the tracking policy is explicitly trained over the reference domain induced by the motion library and bounded modulation parameters, including spatial goals $p_{\mathrm{Goal}}$ used by downstream interaction policies. This exposes the controller during training to the same classes of reference variation that it will encounter when modulation is applied online (Sec.~\ref{sec:generalized_tracking}).  Providing the tracking policy with only the kinematic reference trajectory $\mathbf{q}^{\mathrm{ref}}_t$ rather than the explicit modulation vector $\mathbf{m}_t$ keeps stylistic representation strictly within the generator layer while conditioning the controller to maintain balance across upstream style shifts. In Phase 2, the tracking policy is frozen, and task-specific modulators are trained to map live interaction states to spatial goal adjustments (Sec.~\ref{sec:goal_modulation}). Because these goal modulators act directly on the kinematic reference generator, they require no dynamics model. Consequently, adapting the system to new interaction tasks requires training only a lightweight task modulator rather than retraining the underlying physical tracking policy.

\section{TRACKING OF EXPRESSIVE MOTION}
\label{sec:generalized_tracking}
    
    \subsection{Reference Domain}
    \label{sec:domain}
        
        During deployment, the character executes a continuously generated reference stream $\mathbf{q}^{\mathrm{ref}}_t$ synthesized by sequencing motion library primitives alongside online parameter modulation $\mathbf{m}_t$ (Eq.~\ref{eq:reference_stream}). Phase convergence ($x \to 0$) drives natural primitive transitions, though they can also be forced mid-execution via interactive interrupts. Modulation parameters ($\mathbf{m}_t$) update across a frequency spectrum ranging from real-time control steps to fixed constants.
        
        To expose the policy to this continuous space during training, we represent the domain through a reference dataset of $M$ finite streams, each a sequence of reference poses of length $L$:
        \begin{equation}
            \mathcal{D}_{\mathrm{ref}} = \left\{ \, \mathbf{q}^{\mathrm{ref}}_{0:L,\,i} \, \right\}_{i=1}^M,
            \label{eq:reference_domain}
        \end{equation}
        with $M = 1024$ streams of $L = 1500$ frames ($30\,\mathrm{s}$); evaluation (Sec.~\ref{sec:exp_tracking}) uses a further $2048$ streams sampled independently over the same domain. Primitives and modulation parameters are drawn uniformly from the library and bounding range $[\mathbf{m}_{\min}, \mathbf{m}_{\max}]$, respectively. The modulation includes random goal shift $p_{\mathrm{Goal}}$ to cover non-stationary deployment variations and align $\mathcal{D}_{\mathrm{ref}}$ with downstream task modulation (Sec.~\ref{sec:goal_modulation}). Long stream durations $L$ ensure that each sequence spans multiple randomly sampled primitive transitions and modulation shifts. Finally, training episodes ($T_{\mathrm{ep}} < L$) are sampled from random starting phases $\phi$, adopting Reference State Initialization (RSI)~\cite{peng2018deepmimic} for phase initialization and transferring its concept to mid-stream primitive switches, while preserving sufficient remaining trajectory length.
        
        This domain is defined strictly over the joint articulation space, decoupling physical tracking from expressivity synthesis. Expressive intent is inherited directly from the kinematic DMP reference layer (Sec.~\ref{sec:bg_dmp}), eliminating additional style-seeking reward terms and reducing optimization to joint-space tracking under robot dynamics. Consequently, root trajectory specification remains optional and decoupled from the expressive domain.

    \subsection{Tracking Policy}
    \label{sec:tracking_policy}
        
        The tracking policy $\pi_\theta$ executes physical control over domain $\mathcal{D}_{\mathrm{ref}}$ at $\Delta t = 0.02\,\mathrm{s}$ ($50\,\mathrm{Hz}$). An episode draws a stream from $\mathcal{D}_{\mathrm{ref}}$ together with a random initial phase offset $\phi$, initializing the robot at reference pose $\mathbf{q}^{\mathrm{ref}}_\phi$ and velocity $\dot{\mathbf{q}}^{\mathrm{ref}}_\phi$. Dynamics are randomized per episode over rigid-body properties (base mass, inertia, center of mass), contact friction, and actuation (PD gains, joint damping, armature), collected in the randomization state $\boldsymbol{\zeta}$.
        
        \subsubsection{Observation Space}
            To support sim-to-real deployment, actor observations $\mathbf{o}_t \in \mathbb{R}^{6 + (6+K)n}$ are restricted to onboard-measurable quantities:
            \begin{equation}
                \begin{split}
                    \mathbf{o}_t = \big[\, &\mathbf{g}_b,\ \boldsymbol{\omega}_b,\ \mathbf{q} - \mathbf{q}_0,\ \dot{\mathbf{q}},\ \mathbf{a}_{t-1},\ \mathbf{a}_{t-2}, \\
                    &\mathbf{q}^{\mathrm{ref}}_{\phi+t} - \mathbf{q}_0,\ \dot{\mathbf{q}}^{\mathrm{ref}}_{\phi+t},\ \{ \mathbf{q}^{\mathrm{ref}}_{\phi + t + \Delta_k} - \mathbf{q}_0 \}_{k=1}^{K} \,\big]^\top + \boldsymbol{\varepsilon}_t,
                \end{split}
                \label{eq:actor_obs}
            \end{equation}
            where $\mathbf{g}_b$ is projected gravity, $\boldsymbol{\omega}_b$ is body angular velocity, $\mathbf{q}, \dot{\mathbf{q}}$ are joint states, $\mathbf{q}_0$ is default pose, $\mathbf{a}_{t-1}, \mathbf{a}_{t-2}$ are prior actions, $\{\Delta_k\}_{k=1}^K$ are lookahead offsets, and $\boldsymbol{\varepsilon}_t$ models sensor noise (zeroed on action-history and reference terms).
            The privileged critic network $V_\psi(\mathbf{o}^{\mathrm{c}}_t)$ receives physical state $\mathbf{o}^{\mathrm{c}}_t$ for variance reduction: $\mathbf{o}^{\mathrm{c}}_t = [\mathbf{o}^{\text{clean}}_t, \mathbf{v}_b, z, \{\mathbf{p}^{\mathrm{c}}_i - \mathbf{p}^{\mathrm{root}}\}_i, \boldsymbol{\kappa}, \boldsymbol{\zeta}]^\top$, incorporating clean actor terms alongside true base linear velocity $\mathbf{v}_b$, base height $z$, root-relative contact positions $\mathbf{p}^{\mathrm{c}}_i - \mathbf{p}^{\mathrm{root}}$, contact mask $\boldsymbol{\kappa}$, and randomization state $\boldsymbol{\zeta}$.
        
        \subsubsection{Action Space}
            The policy outputs residual position offsets $\mathbf{a}_t \in [-1, 1]^n$ relative to current reference pose $\mathbf{q}^{\mathrm{ref}}_{\phi+t}$. Commanded joint positions are clipped to joint limits $[\underline{\mathbf{q}}, \overline{\mathbf{q}}]$ and tracked by a PD servo yielding motor torques $\boldsymbol{\tau}_t$:
            \begin{equation}
                \mathbf{q}^{\mathrm{cmd}}_t = \operatorname{clip}\big(\mathbf{q}^{\mathrm{ref}}_{\phi+t} + \eta \mathbf{a}_t, \, \underline{\mathbf{q}}, \, \overline{\mathbf{q}}\big),
                \label{eq:action}
            \end{equation}
            where $\eta = 0.5$ rad scales residual authority. Target clipping ensures safe commands while preserving the unclipped reference in the observation.
        
        \subsubsection{Reward Formulation}
        The scalar reward $r_t$ balances imitation tracking against stability, smoothness, and safety constraints:
        \begin{equation}
            r_t = r^{\mathrm{imit}}_t + r^{\mathrm{vel}}_t + w_{\mathrm{alive}} - r^{\mathrm{anc}}_t - r^{\mathrm{slip}}_t - r^{\mathrm{smooth}}_t - r^{\mathrm{lim}}_t.
            \label{eq:reward}
        \end{equation}
        To prevent unearnable rewards when expressive modulations exceed joint limits, position tracking is evaluated against the feasibility-projected reference $\hat{\mathbf{q}}^{\mathrm{ref}}_{\phi+t} = \operatorname{clip}(\mathbf{q}^{\mathrm{ref}}_{\phi+t}, \underline{\mathbf{q}}, \overline{\mathbf{q}})$:
        \begin{equation}
            r^{\mathrm{imit}}_t = w_{\mathrm{pos}} \exp\!\left( - \frac{k_{\mathrm{pos}}}{n} \sum_{j=1}^n \lambda_j \big( q_j - \hat{q}^{\mathrm{ref}}_{\phi+t,j} \big)^2 \right),
            \label{eq:r_imit}
        \end{equation}
        and velocity matching takes the same exponential form over joint velocity errors $\dot{q}_j - \dot{q}^{\mathrm{ref}}_{\phi+t,j}$ with sensitivity $k_{\mathrm{vel}}$ and weight $w_{\mathrm{vel}}$. Normalization by joint count $n$ makes the sensitivities morphology-invariant.  The mask $\lambda_j$ 
        relaxes imitation on humanoid leg joints (0.1 vs. 1.0 elsewhere) and is 1.0 on Go2 ~\cite{cheng2024expressive,he2024h2o}.
        Survival bonus $w_{\mathrm{alive}}$ provides a dense incentive to maintain stability.
        
        Keeping an in-place gesture in place must not forbid the excursions the gesture itself calls for. The anchor term therefore penalizes not the instantaneous root-pose error but its exponential moving average $\overline{(\cdot)}$, taken with a time constant $T_{\mathrm{anc}}$ that only penalizes drifts across longer execution:
        \begin{equation}
            r^{\mathrm{anc}}_t = w^{\mathrm{pos}}_{\mathrm{anc}} \big\| \overline{\mathbf{p}_{xy} - \mathbf{p}^{\mathrm{target}}_{xy}} \big\| + w^{\mathrm{yaw}}_{\mathrm{anc}} \big| \overline{\psi_{\mathrm{st}} - \psi^{\mathrm{target}}} \big|,
            \label{eq:r_anchor}
        \end{equation}
        where $\mathbf{p}_{xy}$ is the base position and $\psi_{\mathrm{st}}$ the stance heading, which is the least-squares planar rotation between the currently grounded contact points and their layout in the default pose. Restricting the fit to grounded contacts decouples body heading from upper-body gestures, raised limbs, and any reorientation achieved without moving the feet. As a result, actions like aiming and closed-loop heading control (Sec.~\ref{sec:goal_modulation}) do not register as body turns unless they force an actual change in foothold placement, offering a mechanism to preserve primitive expressivity that also keeps this term compatible with interactive goal modulation.

        The remaining terms are regularizers with penalties to avoid foot skating $r^{\mathrm{slip}}_t$ ($w_{\mathrm{slip}}$), high first and second differences of the action as well as high joint torque and acceleration $r^{\mathrm{smooth}}_t$ ($w_{\mathrm{ar}}$, $w_{\mathrm{aa}}$, $w_{\tau}$, $w_{\ddot{q}}$), and incursion into a soft margin $\Delta q$ inside the joint limits $r^{\mathrm{lim}}_t$ ($w_{\mathrm{lim}}$).

        An episode terminates when any body other than a designated contact body touches the ground. With total step rewards weighted positively while upright, this creates an incentive to maintain stability in addition to the dense reward.

\section{INTERACTIVE GOAL MODULATION}
\label{sec:goal_modulation}

    Interactive behavior is realized by modulating only the DMP goal $g$ with $p_{\mathrm{Goal}}$, leaving the tracking objective unchanged. The resulting reference is then executed by the frozen tracking policy in a closed loop. This adaptation propagates through a closed-loop control system: upon a given interaction state, a goal shift modifies the reference trajectory, which the fixed tracking policy (Sec.~\ref{sec:generalized_tracking}) executes, thereby again altering the interaction state that informs subsequent goal updates. Consequently, the goal-modulation policy is learned end-to-end within this closed loop against a frozen tracking controller.
    
    Each interaction task is governed by a lightweight policy, $\pi^{\mathrm{mod}}_{\omega}\!\left( \Delta \mathbf{g}_t \mid \mathbf{o}^{\mathrm{int}}_t \right)$,
    which maps the current interaction observation $\mathbf{o}^{\mathrm{int}}_t$ to a per-joint offset $\Delta \mathbf{g}_t$ on the goal $g$ of the reference-generating DMP at control frequency. The observation vector encodes the relative interaction state, such as target coordinates within the robot's task space.
    
    The generated offset is bounded element-wise via
    \begin{equation}
        \Delta\mathbf{g}_t = \delta \tanh f_\omega(\mathbf{o}^{\mathrm{int}}_t),
        \label{eq:modulator_out}
    \end{equation}
    where $f_\omega$ denotes the neural network parameterized by $\omega$, and $\delta$ represents the maximum permissible shift. The interface itself admits analytic or otherwise hand-designed mechanisms in its place. To maintain tracking performance, $\delta$ is constrained within the goal-modulation range used during tracking-policy training. Additionally, spatial modulation can be restricted to target degrees of freedom via a task-specific joint mask, isolating the kinematic chains that contribute to the interaction, e.g., an arm for a pointing task.
    
    The interaction task is specified by the reward function
    \begin{equation}
        r_t^{\mathrm{int}} = w_{\mathrm{task}} \exp\!\Big(-k \left\| \mathbf{e}^{\mathrm{task}}_t \right\|^2\Big) - r^{\mathrm{mag}}_t - r^{\mathrm{rate}}_t,
        \label{eq:modulator_reward}
    \end{equation}
    where $\mathbf{e}^{\mathrm{task}}_t$ measures the task error evaluated on the interaction state. The magnitude penalty
    \begin{equation}
        r^{\mathrm{mag}}_t = w_{\mathrm{mag}} \big\| \Delta\mathbf{g}_t \big\|^2
        \label{eq:modulator_mag}
    \end{equation}
    regularizes goal displacements to the minimum required for task execution, preserving unmodulated baseline movements and non-essential joints. To maintain smoothness and mitigate oscillations around stationary target states, we penalize high-frequency adaptation
    \begin{equation}
        r^{\mathrm{rate}}_t = w_{\mathrm{rate}} \big\| \Delta\mathbf{g}_t - \Delta\mathbf{g}_{t-1} \big\|^2 .
        \label{eq:modulator_rate}
    \end{equation}
    
    The proposed decoupled architecture offers key system-level advantages. Isolating the tracking controller from task objectives avoids introducing extra complexity into the tracking policy objective. Furthermore, the lightweight parameterization of the modulator enables sample-efficient learning and allows task-specific modules to be swapped at runtime in a modular, plug-and-play fashion.

\section{IMPLEMENTATION AND TRAINING}
\label{sec:implementation}

    All training is conducted in the Newton physics engine (v1.2.1) on a single NVIDIA RTX 5090, with PD gains and $\lambda_j$ set per character. Settings for the optimization and training setup are given in Table~\ref{tab:setup}.
    The reference domain $\mathcal{D}_{\mathrm{ref}}$ is synthesized offline for rollout efficiency in the parallelized setting. Each stream switches primitive and modulation several times within it, drawing $p_{\mathrm{Exa}}, p_{\mathrm{Time}} \in [0.5, 2.0]$, $p_{\mathrm{Arc}} \in [-3, 3]$, $p_{\mathrm{Ant}}, p_{\mathrm{Follow}} \in [0, 0.5]$ and a per-dimension goal shift $p_{\mathrm{Goal}} \in [-0.5, 0.5]\,\mathrm{rad}$.
    The training of interactive goal modulation cannot use a precomputed domain, since $\Delta\mathbf{g}_t$ makes the reference state-dependent. The DMP is therefore integrated into the environment step and batched on the GPU, and unless a fall occurs, new episodes are started without resetting the environment, leaving the robot and DMP phase uninterrupted.

    \subsection{Sim-to-Real Configuration}
    \label{sec:sim2real}

        \begin{table}[t]
        \vspace{0.25cm}
            \centering
            \caption{Training setup. Spanning entries are shared by the tracking and the goal-modulation environment.}
             \vspace{-0.25cm}
            \label{tab:setup}
            \setlength{\tabcolsep}{5pt}
            \footnotesize
            \begin{tabular}{l cc}
                \toprule
                                                    & \textbf{Tracking}              & \textbf{Modulation} \\
                \midrule
                \multicolumn{3}{l}{\textbf{Environment}} \\
                Control / physics step              & \multicolumn{2}{c}{$0.02$ / $0.005\,\mathrm{s}$} \\
                Horizon             & \multicolumn{2}{c}{$1000$ steps} \\
                Mass/inertia scale, CoM             & \multicolumn{2}{c}{$0.9$--$1.15$, $\leq 2\,\mathrm{cm}$} \\
                Friction/gain/damping scale         & \multicolumn{2}{c}{$\pm 10$--$20\,\%$} \\
                Environments $N$                    & $4096$                & $256$ \\
                Lookahead $\{\Delta_k\}$, $\eta$    & $\{5,10,15\}$, $0.5$  & -- \\
                \midrule
                \multicolumn{3}{l}{\textbf{Optimization} (PPO)} \\
                Rollout/epochs/minibatches          & \multicolumn{2}{c}{$24$ / $5$ / $4$} \\
                GAE $\lambda$, clip $\epsilon$      & \multicolumn{2}{c}{$0.95$, $0.2$} \\
                Discount $\gamma$                   & $0.997$               & $0.99$ \\
                Learning rate                       & $2\!\times\!10^{-4}$  & $1\!\times\!10^{-3}$ \\
                \quad adaptation                    & KL $0.008$            & fixed \\
                Exploration                         & entropy $0.005$       & $\sigma = 0.15$ \\
                Hidden layers (ELU)                 & $[1024,512,256]$      & $[64,64]$ \\
                \midrule
                \multicolumn{3}{l}{\textbf{Tracking reward}} \\
                $w_{\mathrm{pos}}$, $k_{\mathrm{pos}}$ & \multicolumn{2}{c}{$1.2$, $24$} \\
                $w_{\mathrm{vel}}$, $k_{\mathrm{vel}}$ & \multicolumn{2}{c}{$0.05$, $0.24$} \\
                $w_{\mathrm{alive}}$, $w_{\mathrm{slip}}$ & \multicolumn{2}{c}{$0.05$, $0.2$} \\
                $w^{\mathrm{pos}}_{\mathrm{anc}}$, $w^{\mathrm{yaw}}_{\mathrm{anc}}$, $T_{\mathrm{anc}}$ & \multicolumn{2}{c}{$0.15$, $0.1$, $2.0\,\mathrm{s}$} \\
                $w_{\mathrm{lim}}$, $\Delta q$      & \multicolumn{2}{c}{$0.1$, $0.1\,\mathrm{rad}$} \\
                $w_{\mathrm{ar}}$, $w_{\mathrm{aa}}$ & \multicolumn{2}{c}{$1.7\!\times\!10^{-3}$, $8.7\!\times\!10^{-4}$} \\
                $w_{\tau}$, $w_{\ddot{q}}$          & \multicolumn{2}{c}{$3.5\!\times\!10^{-6}$, $3.5\!\times\!10^{-8}$} \\
                \bottomrule
            \end{tabular}
        \end{table}

        Transfer to a physical platform introduces two sources of mismatch that are absent from the simulated characters: actuator saturation and system latency. Both are modeled inside the training environment rather than compensated for at runtime. Torque is constrained to the platform's per-joint limits within the physics solver, so the constraint is enforced at every substep. Latency is applied as an integer step delay, $0$--$2$ steps on the proprioceptive observation and $0$--$1$ step on the commanded targets, resampled per environment to obtain robustness over this range. The reference window and the action history are excluded, being computed onboard within the same step. Domain randomization is broadened and extended to include pushes of up to $1.0\,\mathrm{m/s}$ at random intervals.

\section{EXPERIMENTS AND RESULTS}
\label{sec:experiments}

    We evaluate the system across three morphologically distinct platforms to address three core questions. First, we investigate whether a single tracking policy per platform reliably executes the specified bounded modulation domain (Sec.~\ref{sec:exp_tracking}). Second, we assess whether goal modulators generate reactive behaviors when layered over the frozen tracker (Sec.~\ref{sec:exp_interaction}). Finally, we evaluate qualitative performance across both simulation and physical hardware (Sec.~\ref{sec:exp_qualitative}).

    \subsection{Platforms and Interaction Tasks}
    \label{sec:exp_setup}
    
        We evaluate our framework across a 12-DoF Unitree Go2 quadruped, a 20-DoF ROBOTIS OP3 humanoid, and a 29-DoF Unitree G1 humanoid. Each platform incorporates an in-place Dynamic Movement Primitive (DMP) gesture library derived either from native keyframes (Go2, OP3) or retargeted human motion capture trajectories (G1). The tracking policy was trained for 10,000 iterations, taking between 1 hour (Go2) and 3 hours (G1) on a single GPU.

        To evaluate task-level modulation, we instantiate interaction tasks with a simulated 3D target point continuously moving between waypoints. The interaction observation vector encodes the 3D target direction in the robot frame $\mathbf{o}^{\mathrm{int}}_t = \hat{\mathbf{v}}_t$,
        where normalized active primitive phase is concatenated to $\mathbf{o}^{\mathrm{int}}_t$ when phase-gating is active.
        
        We train the goal modulator (Sec.~\ref{sec:goal_modulation}) on two tasks, \emph{Facing} and \emph{Pointing}, which both minimize task error
        \begin{equation}
            e^{\mathrm{task}}_t = \arccos\!\big( \hat{\mathbf{d}}_t^\top \hat{\mathbf{v}}_t \big),
            \qquad
            \hat{\mathbf{d}}_t = \frac{\mathbf{d}_t}{\lVert \mathbf{d}_t \rVert},
            \label{eq:task_error}
        \end{equation}
        \emph{Facing} aligns heading $\mathbf{d}_t$ (forward torso for Go2, head for OP3, chest normal for G1) across all library primitives. \emph{Pointing} aligns an end-effector vector $\mathbf{d}_t$ (base link $\mathbf{b}_t$ to tip $\mathbf{h}_t$) selectively on directional primitives. For gestures starting and ending near default joint states, modulation is restricted to a central phase window to preserve boundaries.
        To prevent non-pointing limb postures, pointing includes an extension term $w_{\mathrm{ext}} \min(\lVert \mathbf{h}_t - \mathbf{b}_t \rVert / R, 1)$ ($R$ represents maximum reach). Modulated degrees of freedom are limited to task-relevant kinematic chains (e.g., 12 DoFs for Go2 base steering vs.\ arm joints for G1). Each policy was trained for 1000 iterations ($<10$\,min/task) with individually tuned reward weights.

    \subsection{Metrics}
    \label{sec:exp_metrics}
    
        Physical execution is evaluated using episode survival rate (Surv.) and reference-tracking error relative to feasibility-projected reference configurations. Reference-tracking fidelity is quantified using root-relative Mean Per-Joint Position Error (MPJPE) over forward-kinematic body positions, one keypoint per actuated joint, and the Mean Absolute Error (MAE) of joint angles. Root-relative Cartesian error is independent of joint count, while evaluating against modulated references captures attenuation of the commanded motion. Global evaluation is infeasible because some modulations are inherently unstable. Errors are computed over the streams completed by all execution variants. This subset is dominated by mild modulations and can therefore be biased in favor of narrow-domain policies. Survival is the primary criterion and puts this into perspective.
        Interaction tasks are evaluated using the averaged task error $e_t^{\mathrm{task}}$ in degrees (Aim) as well as the introduced execution related metrics. Phase-activated modulation is evaluated on the central interval of the primitive, to not include deliberately designed start and end at the default pose, as errors in the evaluation.

    \subsection{Tracking Performance and Execution Stability}
    \label{sec:exp_tracking}

        \begin{table*}[t]
        \vspace{0.3cm}
\centering
\caption{Quantitative tracking and interaction results. Arrows ($\uparrow,\downarrow$) indicate the direction of optimal performance.}
             \vspace{-0.25cm}
\label{tab:results}
\footnotesize
\setlength{\tabcolsep}{2.5pt}
\renewcommand{\arraystretch}{1.0}
\newcommand{\rot}[1]{\rotatebox[origin=c]{90}{\scriptsize\textbf{#1}}}
\newcommand{\hdr}[2]{#1~\tiny#2}
\begin{tabular}{@{}c@{\hspace{5pt}}l@{\hspace{7pt}}cccc@{\hspace{7pt}}cccc@{\hspace{7pt}}cccc@{}}
\toprule
& & \multicolumn{4}{c}{\textbf{Go2}} & \multicolumn{4}{c}{\textbf{OP3}} & \multicolumn{4}{c}{\textbf{G1}} \\
\cmidrule(lr){3-6}\cmidrule(lr){7-10}\cmidrule(lr){11-14}
& & \hdr{Surv.}{\%$\uparrow$} & \hdr{MPJPE}{mm$\downarrow$} & \hdr{MAE}{$^\circ\downarrow$} & \hdr{Aim}{$^\circ\downarrow$}
& \hdr{Surv.}{\%$\uparrow$} & \hdr{MPJPE}{mm$\downarrow$} & \hdr{MAE}{$^\circ\downarrow$} & \hdr{Aim}{$^\circ\downarrow$}
& \hdr{Surv.}{\%$\uparrow$} & \hdr{MPJPE}{mm$\downarrow$} & \hdr{MAE}{$^\circ\downarrow$} & \hdr{Aim}{$^\circ\downarrow$} \\
\midrule
\multirow{5}{*}{\rot{Tracking}}
& \textit{RoboMMT} & $\mathbf{99.66}$ & $3.73$ & $1.81$ & --- & $\mathbf{99.37}$ & $12.65$ & $4.24$ & --- & $\mathbf{99.32}$ & $24.11$ & $5.29$ & --- \\
& \quad\textit{w/o} $\mathcal{D}_\mathrm{ref}$ & $70.12$ & $3.31$ & $1.32$ & --- & $65.43$ & $4.58$ & $1.94$ & --- & $49.95$ & $27.13$ & $5.75$ & --- \\
& \quad\textit{Ref.-stream} & $74.12$ & --- & --- & --- & $24.27$ & --- & --- & --- & $0.00$ & --- & --- & --- \\
& \textit{GMT}~\cite{chen2025gmt} & --- & --- & --- & --- & --- & --- & --- & --- & $83.59$ & $25.65$ & $5.71$ & --- \\
& \textit{SONIC}~\cite{luo2025sonic} & --- & --- & --- & --- & --- & --- & --- & --- & $95.02$ & $74.53$ & $11.77$ & --- \\
\midrule
\multirow{3}{*}{\rot{Facing}}
& \textit{RoboMMT} & $100.00$ & $5.80$ & $2.52$ & $6.60$ & $100.00$ & $12.14$ & $4.19$ & $17.67$ & $100.00$ & $23.78$ & $5.31$ & $17.42$ \\
& \quad\textit{w/o modulation} & $100.00$ & $3.56$ & $1.78$ & $17.04$ & $100.00$ & $12.09$ & $4.23$ & $28.16$ & $100.00$ & $23.37$ & $5.32$ & $27.93$ \\
& \quad\textit{w/o} $\mathcal{D}_\mathrm{ref}$ & $100.00$ & $7.10$ & $3.18$ & $8.50$ & $100.00$ & $3.90$ & $1.74$ & $18.30$ & $100.00$ & $29.40$ & $6.04$ & $19.90$ \\
\midrule
\multirow{3}{*}{\rot{Pointing}}
& \textit{RoboMMT} & $100.00$ & $9.22$ & $4.44$ & $9.69$ & $100.00$ & $10.76$ & $3.34$ & $6.20$ & $100.00$ & $9.97$ & $2.72$ & $7.01$ \\
& \quad\textit{w/o modulation} & $100.00$ & $2.25$ & $1.37$ & $105.95$ & $100.00$ & $10.73$ & $4.09$ & $98.53$ & $100.00$ & $9.52$ & $2.71$ & $20.00$ \\
& \quad\textit{w/o} $\mathcal{D}_\mathrm{ref}$ & $98.00$ & $8.40$ & $4.81$ & $42.10$ & $100.00$ & $4.40$ & $1.59$ & $7.30$ & $100.00$ & $13.60$ & $2.47$ & $9.00$ \\
\bottomrule
\end{tabular}
\end{table*}
    
        \begin{figure}[t]
            \vspace{0.3cm}
            \centering\includegraphics[width=0.95\linewidth]{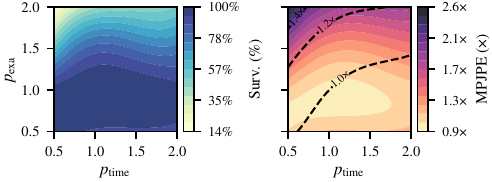}
             \vspace{-0.3cm}
            \caption{Modulation-plane sweep ($p_{\text{Exa}}$ vs. $p_{\text{Time}}$), for training \textit{w/o} $\mathcal{D}_\mathrm{ref}$ (averaged over Go2, OP3, G1) showing survival (left) and relative MPJPE normalized to neutral (right; dashed: \textit{RoboMMT}). Errors increase, and survival drops with distance from the training domain, particularly at high-energy execution ($p_{\text{Exa}} \to 2, p_{\text{Time}} \to 0.5$).}
            \label{fig:sweep}
        \end{figure}

        \begin{figure}
            \centering
            \includegraphics[width=0.95\linewidth]{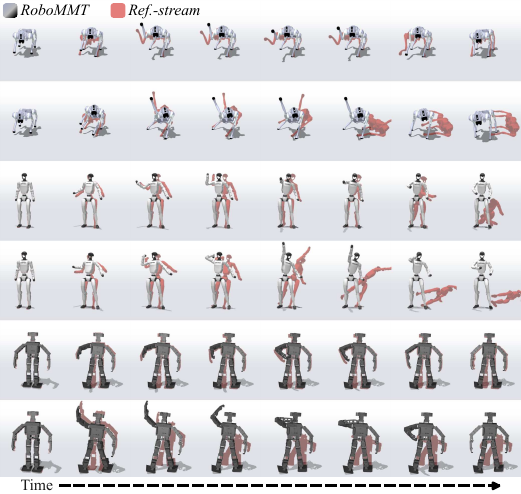}
             
            \caption{Motion sequences showing low-energy (top) and high-energy (bottom) modulations of the same primitive for each character. Compared to the reference stream, \textit{RoboMMT} maintains stable execution for G1 and high energy modulations of the Go2 and OP3.}
            \label{fig:qual-tracker}
        \end{figure}

        Each tracking policy is evaluated on $2048$ independently sampled streams of $500$ steps covering the full modulation domain except $p_{\mathrm{Goal}}$, isolating expressive-style coverage from the goal-directed coverage evaluated in Sec.~\ref{sec:exp_interaction}. All three policies complete $99.3\,\%$ or more of the streams (Table~\ref{tab:results}).
        As a controlled ablation, we train otherwise identical policies only on unmodulated library references. Survival on the modulated evaluation streams then drops to $70.1\,\%$, $65.4\,\%$, and $50.0\,\%$ for Go2, OP3, and G1, respectively. The sweep in Fig.~\ref{fig:sweep} locates failure at high energy modulation, where narrow-domain MPJPE reaches $2.6\times$ its neutral value against $1.4\times$ for ours.
        The survival rate of open-loop execution of the reference stream collapses with increasing morphological stability complexity.
        The narrow-domain tracker achieves lower tracking error on the OP3 and Go2 on the streams it completes, but its substantially lower survival indicates that this precision is achieved at the cost of robustness to highly modulated references. \textit{RoboMMT} instead trades a small increase in tracking error for substantially greater execution coverage and stability. On the G1, our policy leads in both stability and accuracy.
        Two published general-purpose whole-body trackers for G1 are driven by the same stream, in their own deployment stacks with native reference conventions unmodified. \textit{GMT}~\cite{chen2025gmt} and \textit{SONIC}~\cite{luo2025sonic} survive $83.6\,\%$ and $95.0\,\%$ against our $99.3\,\%$, at comparable or higher joint error. Both methods exhibit degradation under variations evaluated here, despite their broader motion-tracking training. This suggests that explicit coverage of the target modulation domain can be important for robust tracking of continuously modulated reference streams.

    \subsection{Interactive Goal Modulation}
    \label{sec:exp_interaction}
    
        Closed-loop goal modulation, evaluated over $256$ randomly sampled environments with $1000$ steps, enables online spatial adaptation while maintaining stable execution across the evaluated facing and pointing tasks (Table~\ref{tab:results}), greatly reducing task errors compared to execution without goal modulation. The same trends are observable when comparing against the ablated tracking policy exposed to the narrow training domain. Here specifically, the complex task formulation for the Go2 \textit{Pointing} stands out ($9.69$ vs $42.10$).
        Task complexity also separates the two trackers. On the humanoid tasks, the narrow-domain tracker reports the lower joint error, as in Sec.~\ref{sec:exp_tracking}. These tasks modulate free upper-body chains and remain close to the unmodulated references it was trained on. Go2 \textit{Pointing} instead shifts joints that carry the robot. Such deviations lie outside its training domain, and it loses balance, producing the only survival below 100\,\% observed under interaction. Tracking error is therefore to be read against task success and task complexity.   

    \subsection{Qualitative Results and Deployment}
    \label{sec:exp_qualitative}

        \begin{figure}
            \centering
            \includegraphics[width=0.95\linewidth]{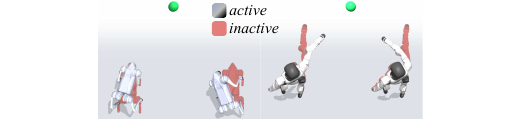}
             
            \caption{Interactive goal modulation in \textit{RoboMMT} enables Go2 to face (left) and G1 to point (right) towards a 3D target point (green).}
            \label{fig:qual-interact}
        \end{figure}

        \begin{figure}[h]
            \centering            \includegraphics[width=0.95\linewidth]{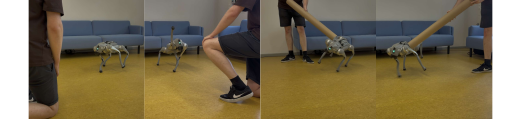}
             
            \caption{Sim-to-real deployment showcasing interactive behavior with a moving interaction partner (left) and stability under disturbance (right) in a long-running modulated sequence.}
            \label{fig:s2r}
        \end{figure}

        With qualitative results in Fig.~\ref{fig:qual-tracker}, we demonstrate that the tracking policy enables stable execution during both low and high energy modulation, where the executed plain reference stream fails. Analyzing the movements reveals that stability is generally induced by shifting body mass or adapting ground contact patterns. For interactive behavior, we show how the robot's pose is adapted to fulfill the spatial interaction task, compared to the plain execution of the reference (Fig.~\ref{fig:qual-interact}). We further deploy both the tracking policy and goal-modulation zero-shot on the physical Go2 using the configuration of Sec.~\ref{sec:sim2real}.
        Qualitatively, we demonstrate modulated interaction sequences with varying expressivity covering the domain, targeted towards a human interaction partner, with the goal modulator toggled on and off during execution. In a representative long-running hardware execution exceeding $90\,\mathrm{s}$, comprising 20 concatenated primitives and externally induced disturbances, the robot completed the sequence without falling across 5 trials.

\section{CONCLUSION}
\label{sec:conclusion}
    We presented \textit{RoboMMT}, a modular architecture for executing expressive body language on dynamically unstable robots. By training a single reference-conditioned tracking policy over the reference domain $\mathcal{D}_{\mathrm{ref}}$ induced by a parameterized DMP motion representation, the controller realizes continuously modulated expressive motion while preserving balance. Layering lightweight goal-modulation policies on top of the frozen tracker adds closed-loop spatial reactivity without altering the tracking objective or requiring a dynamics model. Across a quadruped and two humanoids, the tracker completed at least $99.3\,\%$ of streams sampled from the full modulation domain, whereas an otherwise identical policy trained only on unmodulated references dropped to $50.0$--$70.1\,\%$, with failures concentrated at high-energy modulation. Two general-purpose whole-body trackers driven by the same stream likewise degraded, indicating that explicit coverage of the target modulation domain matters for robust execution. Goal modulators reduced facing and pointing errors substantially at comparable tracking fidelity, and the complete system transferred zero-shot to a physical Go2, sustaining a $90\,\mathrm{s}$ sequence of $20$ concatenated primitives under external disturbances.

\balance
\bibliographystyle{IEEEtran.bst}
\bibliography{references.bib}

\end{document}